\documentclass{article}

\PassOptionsToPackage{numbers}{natbib}
 \usepackage[preprint]{neurips_2026}

\usepackage[utf8]{inputenc} 
\usepackage[T1]{fontenc}    

\usepackage{url}            
\usepackage{booktabs}       
\usepackage{amsfonts}       
\usepackage{nicefrac}       
\usepackage{microtype}      
\usepackage{xcolor}         
\usepackage{bm}
\usepackage{amsmath,amssymb}

\usepackage{algorithm}
\usepackage{algpseudocode}
\usepackage{tikz}
\usetikzlibrary{arrows.meta,positioning,shapes.geometric,fit,calc,decorations.pathreplacing}
\usepackage{pgfplots}
\pgfplotsset{compat=1.18}
\usepackage{subcaption}
\usepackage{enumitem}
\usepackage{tcolorbox}
\usepackage{multirow}
\usepackage{array}
\usepackage{listings}
\usepackage{hyperref}       
\setcitestyle{square}
\setcitestyle{comma}
\usepackage{graphicx} 
\usepackage{colortbl}
\usepackage{wrapfig}
\usepackage{hyperref}       
\usepackage{cleveref}

\hypersetup{
	colorlinks=true,
	citecolor=cyan,
}

\newcommand{\method}{\textsc{TFlow}}

\newcommand{\R}{\mathbb{R}}

\newcommand{\btheta}{\boldsymbol{\theta}}

\newcommand{\bpsi}{\boldsymbol{\psi}}
\newcommand{\bW}{\mathbf{W}}
\newcommand{\bA}{\mathbf{A}}
\newcommand{\bB}{\mathbf{B}}
\newcommand{\bH}{\mathbf{H}}

\newcommand{\bC}{\mathbf{C}}

\newcommand{\bw}{\mathbf{w}}

\newcommand{\bU}{\mathbf{U}}
\newcommand{\bV}{\mathbf{V}}
\newcommand{\bX}{\mathbf{X}}
\newcommand{\bY}{\mathbf{Y}}
\newcommand{\bZ}{\mathbf{Z}}

\newcommand{\bQ}{\mathbf{Q}}
\newcommand{\bE}{\mathbf{T}}
\newcommand{\bS}{\mathbf{T}}

\definecolor{lightgreen}{RGB}{198,239,206}
\definecolor{darkgreen}{RGB}{0,115,0}
\definecolor{pinkred}{RGB}{180,35,35}
\definecolor{impgreen}{RGB}{205,238,205}
\definecolor{goodgreen}{RGB}{0,115,0}
\definecolor{badred}{RGB}{180,35,35}

\newcommand{\impup}[1]{\textbf{\textcolor{darkgreen}{$\uparrow$ #1}}}          
\newcommand{\impdn}[1]{\textcolor{pinkred}{$\downarrow$ #1}}          
\newcommand{\tokimp}[1]{\textbf{\textcolor{darkgreen} {$\downarrow$ #1}}}        

\newcommand{\spdup}[1]{\textbf{\textcolor{darkgreen}{$\times$#1}}}             
\newcommand{\spddn}[1]{\textcolor{pinkred}{$\times$#1}}             

\definecolor{cGreen}  {HTML}{7EC88E}   
\definecolor{cPurple} {HTML}{A88FD0}   
\definecolor{cPink}   {HTML}{DEB0E6}   
\definecolor{cGray}   {HTML}{BEBEBE}   
\definecolor{cBlue}   {HTML}{8AC2EA}   
\definecolor{cTeal}   {HTML}{7AC8BA}   
\definecolor{cOrange} {HTML}{F0AA7A}   
\definecolor{cSalmon} {HTML}{F6CAB2}   

\newcommand{\Hcal}{\mathcal{H}}

\newcommand{\tin}{\text{in}}
\newcommand{\tout}{\text{out}}

\usepackage{booktabs}

\title{Good Agentic Friends Do Not Just Give Verbal Advice: They Can Update Your Weights
}

\author{%
  Wenrui Bao \\
  University of Central Florida \hfill
  \And
  Huan Wang \\
  Westlake University 
  \And
  Jian Wang \\
  Snap Inc.
  \AND
  Zhangyang Wang \\
  UT-Austin
  \And
  Kai Wang \\
  Tencent Hy
  \And
  Yuzhang Shang\thanks{Corresponding author} \\
  University of Central Florida \\
}

\begin{document}

\maketitle
\vspace{-0.3in}
\begin{center}
\href{https://bwr-hhh.github.io/tflow-project-page/}{\textbf{Project}}
~~~~~~
\href{https://github.com/BWR-hhh/TFlow}{\textbf{Code}}
\end{center}
\begin{abstract}

Multi-agent LLM systems usually collaborate by exchanging natural-language messages. This interface is simple and interpretable, but it forces each sender's intermediate computation to be serialized into tokens and then reprocessed by the receiver, thereby increasing the generated-token cost, prefill overhead, and KV-cache memory. We study an alternative communication interface: instead of appending a sender's message to the receiver's context, compile the sender's hidden states into a transient, receiver-specific weight perturbation. We introduce \textbf{\method{}} (Thought Flow), a weight-space communication framework for a known and fixed receiver architecture. For each query, frozen role-prompted sender agents process the input, and a learned parameter generator maps their internal activations into low-rank LoRA perturbations targeting the receiver's modules. These perturbations are fused and applied only during the receiver's generation, enabling instance-level adaptation without permanently changing the model or enlarging the receiver's text context. 
With three Qwen3-4B agents, \textbf{\method{}} improves over a standalone receiver by up to 8.5 accuracy points across five benchmarks while reducing processed tokens by up to 32.69\%. Compared with a text-based three-agent baseline, it reduces total processed tokens by up to 83.27\% and the wall-clock inference time by up to 4.6×, while maintaining competitive accuracy on four of five benchmarks. These results suggest that transient low-rank weight perturbations can serve as an executable communication medium for efficient multi-agent LLM collaboration.
\end{abstract}

\section{Introduction}
\label{sec:introduction}


\begin{figure}[!t]
\centering
\includegraphics[width=1\textwidth]{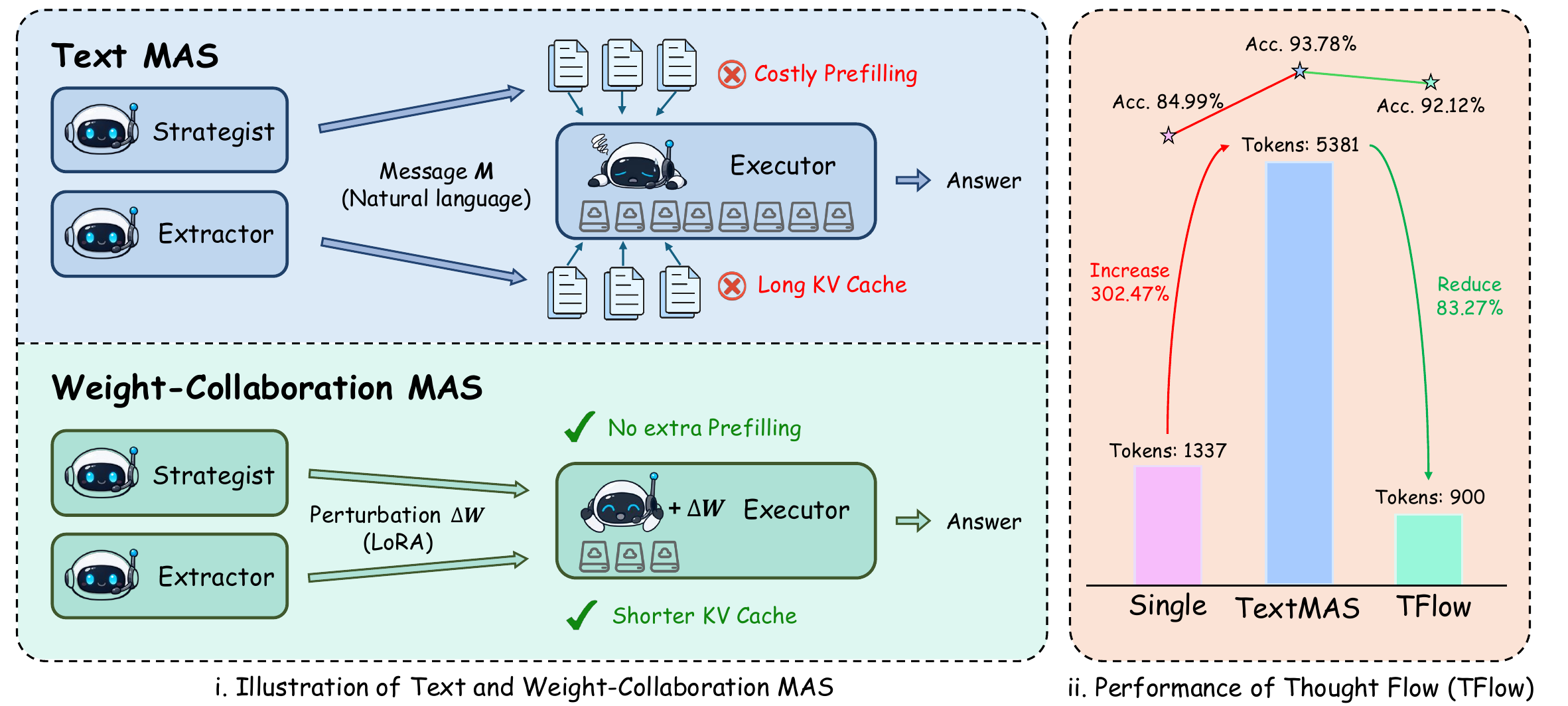}
\caption{\textbf{(i)} Comparison between Text-based MAS and the proposed Weight-Collaboration MAS. In Text MAS, auxiliary agents transmit natural language messages to the Executor, incurring costly prefilling overhead and inflated KV cache. In contrast, our proposed paradigm compresses inter-agent communication into lightweight LoRA weight perturbations $\Delta W$, which are directly merged into the parameters, thereby eliminating the extra prefilling and significantly reducing the KV cache footprint. \textbf{(ii)} Performance overview on \textsc{GSM8K}. \method{} achieves accuracy competitive with TextMAS while reducing token consumption by $\mathbf{76.7\%}$, substantially surpassing the single-agent baseline in both accuracy and efficiency.}
\vspace{-0.3in}
\label{introduction}
\end{figure}

Multi-agent LLM systems are often built like small committees. One model proposes a strategy, another contributes domain knowledge, and a final model produces the answer. This design has been effective across reasoning, coding, and tool-use settings~\citep{li2023camel,hong2023metagpt,wu2024autogen,du2024improving,liang2024encouraging, guo2024large, xi2025rise}. Yet the communication interface between agents has changed surprisingly little: most systems still require each agent to write a natural-language message that is appended to another agent's context.

This text interface is attractive because it is universal and human-readable. \textbf{We yet argue it might be an awkward interface between neural networks}. 
A sender agent has already transformed the input into a rich internal state, but conventional multi-agent systems force this state to be decoded into tokens before the receiver can use it. The receiver must then encode those tokens again, paying additional prefill cost and storing a larger KV cache (Figure~\ref{introduction} (i)), whose memory and latency grow with context length~\citep{pope2023efficiently,kwon2023efficient}. As the number of agents or communication rounds grows, this repeated write-read cycle can dominate the cost of collaboration. The issue is not only that natural-language messages are long; it is that they make model-to-model communication pass through a channel designed for humans.

Recent work has begun to question this design by allowing agents to exchange continuous representations rather than text~\citep{li2021prefix, lester2021power, du2025enabling,zheng2025thought,zou2025latent,jin2026agent}. These latent-space protocols reduce decoding overhead and can preserve information that would be difficult to express in words. However, they still require the receiver to consume sender information as activations, embeddings, or cache states. This creates a different compatibility problem: the transmitted object must be meaningful inside the receiver's representation geometry, making such methods most natural when agents share architectures, hidden spaces, or carefully trained adapters.

We explore a complementary route. \textit{Instead of sending information that the receiver must read, we send information that changes how the receiver computes.} In particular, we ask whether a sender's internal state can be compiled into a temporary low-rank perturbation of the receiver's weights \citep{houlsby2019parameter, hu2022lora}. The receiver then solves the query under this perturbed computation, after which the perturbation is removed, and the frozen base model is restored. In this view, the communication object is not a sentence and not a hidden state injected into the receiver's cache. It is a query-specific update in the receiver's parameter space.

We instantiate this idea in \textbf{\method{}} (Thought Flow), a framework for weight-space communication among LLM agents. We focus on a practical fixed-receiver setting: the receiver architecture and target modules are known, all LLM backbones remain frozen, and only a receiver-specific parameter generator \citep{mahabadi2021parameter, ha2016hypernetworks} is trained. Given a query, each sender agent processes the prompted input once and exposes its hidden states as a conditioning signal. The parameter generator maps this signal into layer- and module-specific LoRA factors for the receiver. Perturbations from multiple senders are fused and injected by transiently patching the forward pass of the receiver's target linear layers. The receiver then generates its answer without seeing any sender-written message in its text context (Figure~\ref{introduction} (ii)).

This design separates \method{} from both text-based and latent-space collaboration. Compared with text-based MAS, \method{} avoids expanding the receiver's input with auxiliary messages, thereby reducing generated-token usage, prefill overhead, and KV-cache growth. Compared with latent-state transfer, \method{} does not ask the receiver to directly interpret a sender's activations. Sender states are instead translated by a learned generator into perturbations that are compatible with the receiver's own weight tensors by construction. The result is an executable communication channel: sender information affects the receiver by modifying the computation path used for the current query.

We evaluate \method{} in a three-agent setting with a frozen Qwen3-4B backbone, where sender agents provide different perspectives through hidden states and the receiver produces the final answer. Across five benchmarks, \method{} improves over the standalone receiver by up to 8.5 accuracy points while reducing total processed tokens by up to 32.69\%. Relative to a text-based three-agent baseline, it cuts processed tokens by up to 83.27\% and wall-clock inference time by up to 4.6x, with competitive accuracy on four of five benchmarks; the remaining gap on HumanEval+ suggests that additional generation budget matters more than inter-agent communication for code synthesis.

Our contributions are:
\begin{itemize}[leftmargin=*]
    \item Motivated by the high communication cost of text-based multi-agent systems, we propose \textbf{\emph{weight-space communication}}, a new communication paradigm for multi-agent LLM collaboration, in which sender information is transmitted as transient low-rank perturbations to a frozen receiver model rather than as text appended to the receiver's context.

    \item Under this paradigm, we introduce \textbf{\method{}}, a receiver-specific framework that transforms sender hidden states into query-specific LoRA factors, integrates multiple sender contributions, and injects the resulting perturbation only during the receiver's generation.

    \item We demonstrate that \method{} achieves substantial efficiency gains across five reasoning, knowledge, and coding benchmarks, dramatically reducing generated-token and latency costs relative to text-based multi-agent collaboration with only negligible performance degradation, while consistently outperforming a standalone receiver.
\end{itemize}
\section{Related Work}

\subsection{LLM-Based Multi-Agent Systems}

\paragraph{Multi-Agent Collaboration Paradigms.}
Large language models have catalyzed a wave of multi-agent systems.
Early explorations such as CAMEL~\citep{li2023camel} and Generative Agents~\citep{park2023generative} demonstrate the power of role-playing dialogue, enabling agents to cooperate through structured natural-language conversations.
Subsequent work applies multi-agent collaboration to complex engineering workflows: MetaGPT~\citep{hong2023metagpt} and ChatDev~\citep{qian2024chatdev} mirror real software companies by assigning specialized roles and routing structured artifacts between them, while AutoGen~\citep{wu2024autogen} and AutoAgents~\citep{chen2023autoagents} provide general-purpose frameworks for orchestrating and automatically generating task-specific agents.
Another work leverages multi-agent debate to improve reasoning quality~\citep{du2024improving,liang2024encouraging}. ReConcile~\citep{chen2024reconcile} organizes round-table conferences among diverse LLM families with confidence-weighted voting.
More recently, Mixture-of-Agents~\citep{wang2025mixtureofagents} and \citep{li2024more} show that iteratively aggregating or even naively scaling diverse model outputs can surpass the strongest individual model.
Despite their architectural diversity, all of the above systems share a common reliance on natural language as the primary medium for inter-agent communication.

\paragraph{Inter-Agent Communication Mechanisms.}
The communication mechanism is central to the effectiveness of any multi-agent system. 
The prevailing paradigm transmits information as natural-language messages concatenated into each receiver's context window~\citep{li2023camel,hong2023metagpt,wu2024autogen,qian2024chatdev}. 
While natural language provides a universal and human-interpretable interface, serializing an agent's intermediate computation into tokens may lose information that is difficult to verbalize and can introduce substantial communication overhead~\citep{pham2024let,zheng2025thought,du2024improving,smit2024should}. 
Recent work has therefore explored non-textual communication channels: CIPHER~\citep{pham2024let} lets agents communicate through probability-weighted output embeddings; Interlat~\citep{du2025enabling} transmits hidden states as a compressed representation of an agent's internal computation; and \citep{zheng2025thought,tang2025augmenting,zou2025latent} further formalize and extend latent-space communication. 
These approaches show that moving beyond natural-language messages can preserve richer intermediate signals and reduce token-level overhead, primarily by communicating in activation or embedding space. 
In this work, we explore a complementary paradigm: \emph{weight-space communication}, where sender-side computation is converted into targeted, instance-specific perturbations of the receiver's model parameters, which guide the receiver's generation.

\subsection{Knowledge Representation and Operations in Weight Space}

\paragraph{Weight-Space Structure and Static Merging.}
A growing body of evidence shows that neural network weight spaces are structured carriers of transferable knowledge~\citep{yang2026model,li2023deep}.
The \emph{task vector} framework~\citep{ilharco2022editing} demonstrates that the element-wise difference between a fine-tuned model and its pretrained initialization encodes the learned task skill.
Building on this insight, a family of model merging methods, including Model Soups~\citep{wortsman2022model}, Fisher merging~\citep{matena2022merging}, TIES-Merging~\citep{yadav2023ties}, and DARE~\citep{yu2024language}, seeks to combine the knowledge of multiple fine-tuned models directly in weight space without additional training.
Despite these advances, all of the above approaches operate in a \emph{static} manner, in which the merging coefficients or perturbation directions are fixed once determined and do not adapt to individual inputs at inference time.

\paragraph{Low-Rank Adaptation and Dynamic Weight Generation.}
LoRAHub~\citep{huang2024lorahub} consists of multiple LoRA modules with learned scalar weights for generalizing the cross-task, while LoRAs composition~\citep{wang2024instance} and LoRA-Flow~\citep{wang2024lora} further enable selection of dynamic adapters at the instance-level and layer-wise, respectively. Going one step further, recent approaches use hypernetworks~\citep{ha2016hypernetworks} to directly \emph{generate} adapter parameters conditioned on input context---from task embeddings~\citep{mahabadi2021parameter}, to few-shot demonstrations~\citep{phang2023hypertuning}, to full documents~\citep{charakorn2026doc}, which eliminates the need for pre-trained adapter libraries \citep{liang2025draganddrop, jin2024conditional, wang2024neural, wang2025scaling, han2026w2t}.

Our framework, \method{}, extends this progression.
Rather than conditioning on static documents or task embeddings, \method{} uses a parameter generator to translate the \emph{dynamic reasoning states} of multiple sender agents into low-rank perturbations of a receiver model's weights.
\begin{figure}[!t]
\centering
\includegraphics[width=1\textwidth]{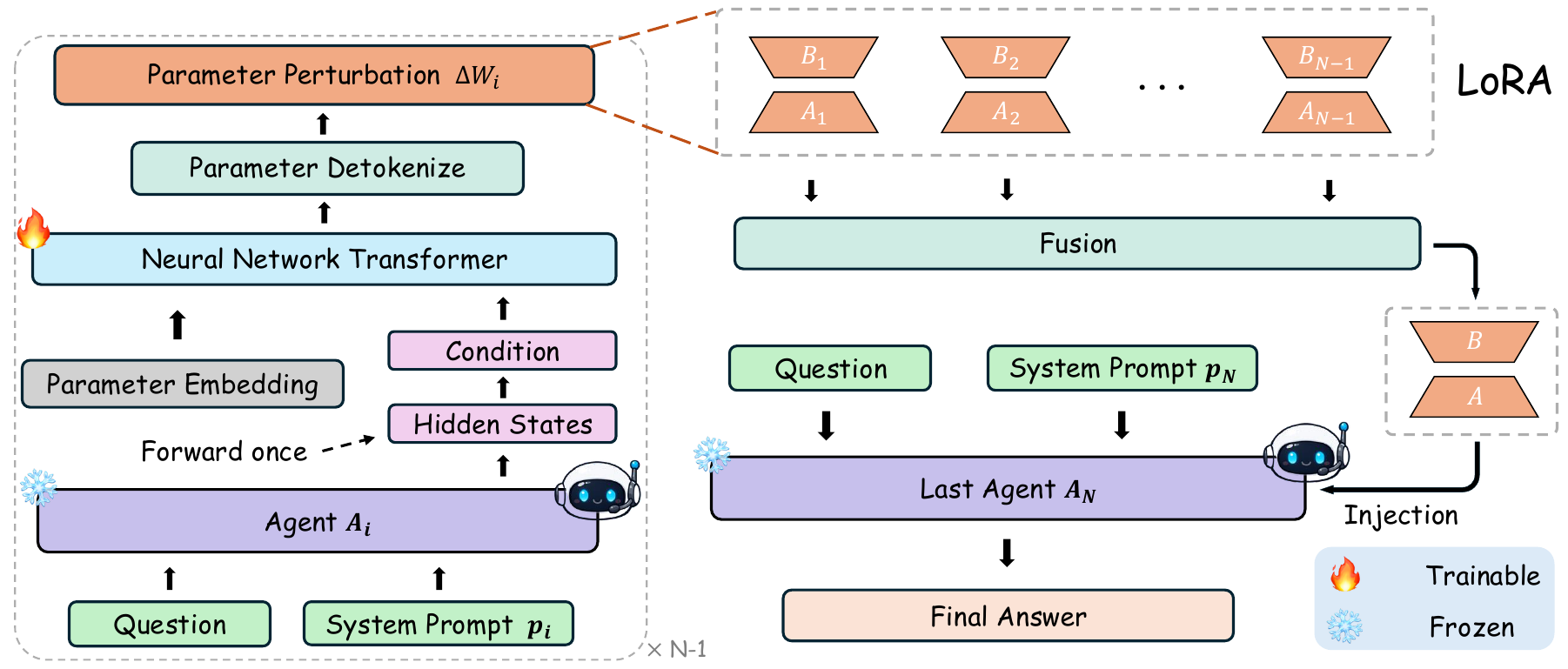}
\caption{
\textbf{Overview of \method{}.}
\method{} realizes multi-agent collaboration through dynamic, instance-specific parameter generation. Given an input question, each sender agent \(A_i\) receives the question together with its system prompt \(p_i\) and performs a single frozen forward pass to produce hidden-state representations that summarize its role-specific reasoning signal. These hidden states serve as conditions for a trainable \textcolor{cBlue}{parameter generator}, which maps them into parameter embeddings, applies a neural transformation, and detokenizes the generated representation into a low-rank perturbation \(\Delta W_i\), parameterized by \textcolor{cOrange}{LoRA factors \((\mathbf{A}_i, \mathbf{B}_i)\)}. The per-sender LoRA factors from all \(N-1\) agents are subsequently combined by a fusion module to form a unified update \(\Delta\bW\). This perturbation is transiently injected into the frozen receiver agent \(A_N\), which is conditioned on the original question and its own system prompt \(p_N\) to generate the final answer. The injected parameters are created on demand for each input and are discarded after generation, leaving the receiver parameters unchanged.
During training, all agents remain \textbf{frozen} and only the parameter generator is \textbf{trainable}.
}
\vspace{-0.212in}
\label{overview}
\end{figure}

\section{\method{}: Thought Flow}
\label{sec:method}

Figure~\ref{overview} illustrates the overall pipeline of \method{}. 
This section formalizes the setup (Section~\ref{sec:method_setup}), describes how sender hidden states are aggregated into query-dependent conditioning signals (Section~\ref{sec:conditioning}), introduces the parameter generator that emits LoRA factors (Section~\ref{sec:generation}), and details the transient injection into the receiver (Section~\ref{sec:fusion_injection}).

\subsection{Problem Setup of Weight-Collaboration MAS}
\label{sec:method_setup}

We consider a collaboration of $N$ agents $\{A_1,\dots,A_N\}$ that share a single frozen backbone $f_{\btheta}$ but are individuated by role-specific system prompts $\{p_1,\dots,p_N\}$. Given a query $q$, agents $A_1,\dots,A_{N-1}$ act as \emph{senders} and $A_N$ as the \emph{receiver} \citep{li2023camel, wu2024autogen, hong2023metagpt}. The senders contribute only their hidden-state trajectories, while the receiver is the one who actually decodes the answer. Rather than exchanging tokens, the senders induce, for this single query, a transient parameter displacement applied to a designated set of linear modules inside the receive:
\begin{equation}
  \hat{\btheta} \;=\; \btheta \;+\; \Delta\bW\!\big(q,\{p_i\}_{i=1}^{N-1};\bpsi\big).
  \label{eq:transient_param}
\end{equation}
Here $\bpsi$ denotes the parameters of a learned generator and is the \emph{only} set of weights updated during training. The backbone tensors themselves are never overwritten: the perturbation is implemented as a temporary forward patch and removed immediately after the receiver call, so different queries always observe the same original backbone.

Two design constraints follow directly and will shape every component below. First, because we rely on the existence of beneficial neighbors rather than on aggressive fine-tuning, $\Delta\bW$ must remain a \emph{low-rank, small-norm} displacement. 
Second, because useful perturbation directions are \emph{instance-specific}, the generator must be conditioned on representations that depend on $q$ itself, not merely on the static role descriptors $p_i$. We therefore condition on the senders' hidden states over the prompted input, which simultaneously encodes the role $p_i$ and the content of the query.

\paragraph{Notation.} In the remainder of this section, we restrict weight-space perturbations to \(L\) selected decoder layers, with \(M\) linear modules perturbed in each selected layer. For the \(m\)-th perturbed module in the \(l\)-th selected layer, we denote its frozen weight matrix by \(\bW_m^{(l)}\).

\subsection{Sender Conditioning}
\label{sec:conditioning}

Each sender \(A_i\) encodes the prompted input \((p_i, q)\) using the frozen backbone in a single forward pass. We retain the intermediate activations and denote the resulting hidden states at layer \(l\) by \(\bH_i^{(l)} \in \R^{T_i \times d}\), for \(l=0,1,\dots,L_{\mathrm{total}}\).

\paragraph{Learnable layer aggregation.}
Instead of relying on a single-layer representation, we aggregate each sender's hidden states across layers using temperature-scaled softmax weights over learnable layer-wise scalars. The resulting conditioning representation \(\bC_i \in \R^{T_i \times d}\) enables the generator to exploit features at different levels of abstraction while preserving a fixed conditioning dimensionality.

\subsection{Parameter Generator}
\label{sec:generation}
The parameter generator \(\Hcal_{\bpsi}\) maps each sender conditioning representation \(\bC_i\) to a complete set of LoRA factors spanning all \(L \times M\) targeted modules in the receiver. The generator comprises three stages: conditioning-driven initialization, a multi-axis Transformer trunk, and detokenization into LoRA matrices.

\paragraph{Stage 1: Initialization.}
We construct a token grid of shape \((L, H \times W, d_{\mathrm{pg}})\), where \(H = H_A + H_B\) divides tokens into \(\bA\)- and \(\bB\)-generating groups, and \(W\) spans the target modules and rank slots. Learnable grid queries \(\bQ_{\mathrm{grid}} \in \R^{L \times HW \times d_{\mathrm{pg}}}\) are initialized by cross-attending to the projected conditioning representation \(\mathrm{Proj}_{\mathrm{in}}(\bC_i)\). We then add layer-index and module-rank positional embeddings to encode each token's structural location within the receiver.

\paragraph{Stage 2: Neural Network Transformer.}
Our generator architecture follows the structured parameter tokenization paradigm introduced by HY-WU~\citep{tencent2026hy}. The initialized grid is refined through \(N_{\mathrm{pg}}\) stacked blocks, each applying three residual attention operations along complementary axes, followed by a feed-forward network. Cross-layer self-attention \((\mathrm{SA}_L)\), batched over the \(HW\) axis, captures dependencies across receiver layers; intra-layer self-attention \((\mathrm{SA}_{HW})\), batched over the \(L\) axis, propagates information across modules and rank slots within each layer; and conditioning cross-attention \((\mathrm{CA})\) re-anchors the generation process to the sender representation. All attention operations employ RoPE~\citep{su2024roformer} along their active sequence axis.

\paragraph{Stage 3: Detokenization.}
After \(N_{\mathrm{pg}}\) blocks, the refined grid is partitioned along the \(H\) dimension into \(\bA\)- and \(\bB\)-tokens. Two linear heads map these tokens to parameter slices, which are rearranged into per-sender LoRA factors \(\bA_{i,m}^{(l)} \in \R^{r \times d_\tin}\) and \(\bB_{i,m}^{(l)} \in \R^{d_\tout \times r}\). The generator is shared across all senders; sender-specific factors are induced solely by the conditioning representation \(\bC_i\).


\subsection{Transient Injection}
\label{sec:fusion_injection}
The generated factors are injected into each targeted module of the receiver as a low-rank weight offset. When multiple senders are active, their generated LoRA updates are fused by a lightweight scalar gate before being applied to the receiver:
\begin{equation}
  \Delta\bW_m^{(l)}=\frac{\alpha}{r}\sum_{i=1}^{N-1}\mathrm{softmax}_i\!\bigl(g_{\bw}(\bC_i)\bigr)\bB_{i,m}^{(l)}\bA_{i,m}^{(l)} .
  \label{eq:lora_inject}
\end{equation}
The resulting \(\Delta\bW_m^{(l)}\) is injected only into the designated receiver module \(m\) at layer \(l\). During the receiver's forward pass, the frozen linear map is evaluated with this transient additive update. After generation, the patch is discarded, ensuring that each input induces its own temporary parameterization and that all subsequent inputs start from the same frozen backbone.

\subsection{Training Objective}

\paragraph{Diversity regulariser.}
To prevent the generator from collapsing toward a query-independent LoRA, we penalise the squared cosine similarity $\mathcal{L}_{\mathrm{div}}(\Theta;x_q) = \cos^{2}\!\bigl(v(x_q),\,v_{\mathrm{prev}}^{(\sigma(x_q))}\bigr)$ between the flattened LoRA vector $v(x_q)$ of the current sample and the cached vector $v_{\mathrm{prev}}^{(\sigma)}$ from the previous step. This loss is minimised at orthogonality, equally penalises parallel and anti-parallel collapse, and incurs no additional forward cost since the cache is updated via stop-gradient after each backward pass.

\paragraph{End-to-end optimization.}
The task loss $\mathcal{L}_{\mathrm{task}}$ is the standard masked next-token cross-entropy computed only over completion tokens, with the receiver conditioned on both the chat-templated prompt and the injected $\Delta W$. The full training objective combines it with the diversity regulariser:
\begin{equation}
  \;\mathcal{L}(\Theta)
       \;=\;
       \mathbb{E}_{(x_q,\,t,\,\sigma)\sim\mathcal{D}}
       \!\Bigl[
         \mathcal{L}_{\mathrm{task}}(\Theta;x_q,t)
         \;+\;
         \lambda_{\mathrm{div}}\,
         \mathcal{L}_{\mathrm{div}}(\Theta;x_q)
       \Bigr].\;
  \label{eq:total_loss}
\end{equation}
The trainable parameter set $\Theta$ encompasses all components of the parameter generator $\mathcal{H}_{\bpsi}$, the conditioning projection layers, and the gating head; the receiver and sender backbones share the frozen parameters \(\btheta\) and remain entirely frozen throughout training. We do not add auxiliary weight-norm or inter-sender cosine penalties in our default objective; empirically, the cache-based diversity regularizer is sufficient to encourage instance-level specialization.
\section{Experiments}
\label{sec:experiments}

\subsection{Experimental Setups}
\label{sec:exp_setup}

\paragraph{Datasets and evaluation.}
We evaluate on five benchmarks spanning mathematical reasoning, code generation, and multi-disciplinary knowledge: GSM8K~\citep{cobbe2021training}, MATH~\citep{hendrycks2021measuring}, MMLU~\citep{gema2025we,hendrycks2020measuring}, HumanEval+~\citep{liu2023your,chen2021evaluating}, and MBPP+~\citep{liu2023your,austin2021program}. All benchmarks are evaluated under the zero-shot, chain-of-thought setting with no in-context exemplars. At inference time, all methods use identical decoding hyperparameters, which is temperature $0.6$, top-$p$ $0.95$.

\paragraph{Backbone and baselines.}
All methods share a single frozen \textbf{Qwen3-4B} \citep{yang2025qwen3} backbone, whose parameters are never updated during parameter generator training or inference. We compare against three baselines: (i)~\textbf{Single-Agent}, which directly prompts Qwen3-4B with a task-level system prompt and represents the backbone's intrinsic capability; (ii)~\textbf{TextMAS}, a three-agent text-level collaboration in which Agents~A and B produce strategy and domain-knowledge passages that are concatenated into Agent~C's context, representing conventional multi-agent cooperation.

\paragraph{Parameter generator configuration.}
The system deploys $N\!=\!3$ agents sharing the same backbone instance: Agent~A analyzes reasoning types and plans chain-of-thought structure, Agent~B retrieves domain knowledge and implicit constraints, and Agent~C solves the problem after the generated LoRA perturbations are injected. Specific system prompts are designed for each role. The parameter generator is a lightweight Transformer ($d_{\mathrm{pg}}\!=\!1024$, head dimension~128, $N_{\mathrm{pg}}\!=\!2$ blocks, token dimension~256, dimension-accumulation factor $c\!=\!2$) that produces LoRA factors of rank $r\!=\!4$ with scaling $\alpha\!=\!8$ ($s\!=\!\alpha/r\!=\!2$).

\paragraph{Training details.}
The training data consist of OpenThoughts \citep{guha2025openthoughts}, Sky-T1 \citep{sky-t1-2025}, and KodCode \citep{xu2025kodcode}. We train for one epoch on 32{,}000 samples using a single NVIDIA RTX PRO 6000, which takes approximately 8 hours.

\begin{table}[!t]
\centering
\renewcommand{\arraystretch}{1.25} 
\caption{
\textbf{Main results across five benchmarks.}
For each task, we report accuracy (\textbf{Acc.} $\uparrow$, \%), average total processed tokens (\textbf{Token} $\downarrow$), and end-to-end wall-clock inference time (\textbf{Speed}, in seconds).
Single, TextMAS, and \method{} are shown side by side.
The \textbf{Improvement} columns compare \method{} with each baseline: $\uparrow$/$\downarrow$ denote accuracy increases/decreases in absolute points, $\downarrow$ denotes relative token reduction, and $\times$ denotes the speed ratio.
\textcolor{darkgreen}{Green} indicates favorable changes, while \textcolor{pinkred}{red} indicates unfavorable changes.
}
\label{tab:main}
\resizebox{0.9\textwidth}{!}{%
\begin{tabular}{c|c|cc|c|ll}
\toprule
\multirow{2}{*}{\textbf{Tasks}} 
  & \multirow{2}{*}{\textbf{Metrics}}
  & \multicolumn{2}{c|}{\textit{Baselines}} 
  & \multirow{2}{*}{\method{}} 
  & \multicolumn{2}{c}{\textbf{Improvement}} \\
\cmidrule(lr){3-4} \cmidrule(lr){6-7}
  & & Single & TextMAS & & $\Delta$\,Single & $\Delta$\,TextMAS \\
\hline
\rowcolor{gray!10}
\multicolumn{7}{c}{\textit{General Task}} \\
\hline
\multirow{3}{*}{\textbf{MMLU-Redux}}
  & Acc. $\uparrow$  & 58.99 & 71.50 & 66.97 & \impup{7.98}  & \impdn{4.53} \\
  & Token $\downarrow$ & 1079   & 4825  & 998   & \tokimp{7.51\%}   & \tokimp{79.32\%} \\
  & Speed (s) & 8226  & 36450 & 9784  & \spddn{0.84}  & \spdup{3.73} \\
\hline
\rowcolor{gray!10}
\multicolumn{7}{c}{\textit{Math \& STEM Task}} \\
\hline
\multirow{3}{*}{\textbf{GSM8K}}
  & Acc. $\uparrow$ & 84.99 & 93.78 & 92.12 & \impup{7.13}  & \impdn{1.66} \\
  & Token $\downarrow$ & 1337  & 5381  & 900   & \tokimp{32.69\%}  & \tokimp{83.27\%} \\
  & Speed (s) & 6230  & 27256 & 5953  & \spdup{1.05}  & \spdup{4.58} \\
\hline
\multirow{3}{*}{\textbf{MATH}}
  & Acc. $\uparrow$ & 16.18 & 26.47 & 23.16 & \impup{7.98}  & \impdn{3.31} \\
  & Token $\downarrow$ & 2782  & 8188  & 2242  & \tokimp{19.41\%}  & \tokimp{72.62\%} \\
  & Speed (s) & 1984  & 5213  & 2258  & \spddn{0.88}  & \spdup{2.31} \\
\hline
\rowcolor{gray!10}
\multicolumn{7}{c}{\textit{Coding Task}} \\
\hline
\multirow{3}{*}{\textbf{MBPP+}}
  & Acc. $\uparrow$ & 59.79 & 68.52 & 67.20 & \impup{7.41}  & \impdn{1.32} \\
  & Token $\downarrow$ & 1533  & 5500  & 1301  & \tokimp{15.13\%}  & \tokimp{76.35\%} \\
  & Speed (s)  & 1998  & 5796  & 2395  & \spddn{0.83}  & \spdup{2.42} \\
\hline
\multirow{3}{*}{\textbf{Humaneval+}}
  & Acc. $\uparrow$ & 56.71 & 75.00 & 65.24 & \impup{8.53}  & \impdn{9.76} \\
  & Token $\downarrow$ & 1756  & 5879  & 1662  & \tokimp{5.35\%}  & \tokimp{71.73\%} \\
  & Speed (s) & 872   & 2512  & 1065  & \spddn{0.82}  & \spdup{2.36} \\
\bottomrule
\end{tabular}%
}
\vspace{-0.2in}
\end{table}

\subsection{Main Results}
\label{sec:main_results}

Table~\ref{tab:main} reports accuracy, average total processed token usage, and wall-clock inference time for \method{}, the Single-Agent baseline, and TextMAS across all five benchmarks.

\paragraph{Accuracy improvement over Single-Agent.}
\method{} consistently outperforms the single-agent baseline across all task categories, with accuracy gains of $+7.13$ to $+8.53$ points.
Notably, these gains come with \emph{fewer} total processed tokens on every benchmark, indicating that weight-space perturbations guide the receiver toward more concise reasoning paths rather than verbose exploration.
\method{} incurs higher end-to-end latency on four of the five benchmarks, because applying instance-specific LoRA perturbations to the backbone prevents efficient batched generation.

\paragraph{Efficiency advantage over TextMAS.}
As shown in Table~\ref{tab:main}, total processed-token consumption is cut by $71\text{--}83\%$ across all five benchmarks, translating to $2.3\text{--}4.6\times$ wall-clock speed-ups.
These gains stem from weight-space injection, entirely bypassing the lengthy prefilling and KV-cache overhead inherent in text-based context concatenation. Compared with TextMAS, the accuracy gap remains modest---within $1.3\text{--}4.5$ points on four of the five tasks.
The sole outlier is HumanEval+ ($\Delta=9.76$).
We attribute this to TextMAS's concatenated context serving a dual role: beyond transferring inter-agent knowledge, it implicitly extends the generation budget, enabling longer outputs that facilitate structural elaboration and self-debugging.
\method{} successfully transfers collaborative knowledge through the weight space but does not alter the output distribution's length characteristics, a distinction that surfaces most clearly in code generation.

\subsection{Instance-level \method{} Analysis}
\label{sec:instance_conditional}

A core property of \method{} is that the LoRA consumed by the receiver is generated \emph{per input}, rather than learned once and reused.
We verify this by examining two points of the pipeline: the conditioning vector~$\mathbf{c}$ that drives the parameter generator (\cref{sec:instance_conditional_hidden}), and the resulting LoRA tensors for the receiver (\cref{sec:instance_conditional_lora}).

\subsubsection{Hidden-State Evidence}
\label{sec:instance_conditional_hidden}

We extract per-layer hidden states for 20 instances per dataset and report mean pairwise cosine similarity in three regimes: within-\textsc{gsm8k}, within-\textsc{mbpp+}, and cross-task (\textsc{gsm8k} $\times$ \textsc{mbpp+}).

\begin{wrapfigure}{r}{0.55\textwidth}
    \vspace{-1.5em}
    \centering
    \includegraphics[width=1\linewidth]{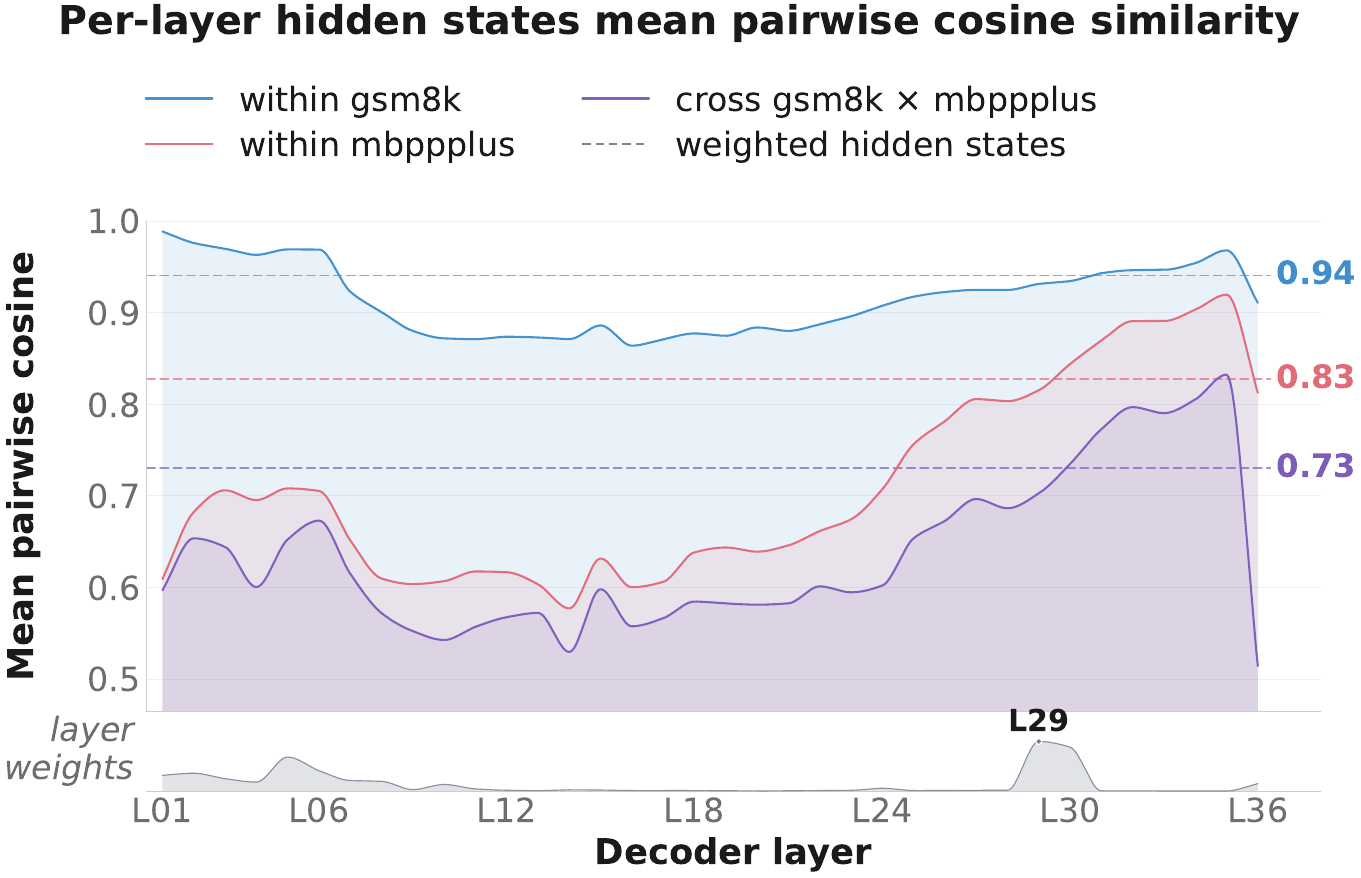} 
    \vspace{-1.6em}
    \caption{Per-layer mean \textbf{pairwise cosine similarity} of the question's last-token hidden state across decoder layers. Solid curves report within-task similarity and cross-task similarity, respectively. Dashed lines indicate the corresponding similarity computed on the aggregated conditioning vector $\mathbf{c}$ that the parameter generator actually consumes. The bottom panel shows the learned layer weights $\{w_\ell\}$, which peak sharply at layer~29.}
    \label{fig:stepgaps}
    \vspace{-2.5em}
\end{wrapfigure}

\paragraph{Learned layer weights.}
As shown in Figure~\ref{fig:stepgaps}, the learned weights $\{w_\ell\}$ do \emph{not} simply track the lowest pairwise-similarity layers.
More than $80\%$ of the mass concentrates on a small set of layers (\textsc{L}05, \textsc{L}29, \textsc{L}30), each with a distinct similarity profile.
This selection emerges purely from end-to-end training, suggesting that these layers are chosen for their conditioning utility rather than surface-level diversity.

\paragraph{Pairwise similarity across regimes.}

Figure~\ref{fig:stepgaps} further shows that the aggregated conditioning vector preserves the expected ordering of semantic similarity: within-GSM8K pairs (0.94) are most similar, followed by within-MBPP+ pairs (0.83), with cross-task pairs (0.73) being the least similar. This confirms that the conditioning representation encodes instance-level variation in a structured manner.

\subsubsection{LoRA Tensor Evidence}
\label{sec:instance_conditional_lora}

For each instance, we materialise the effective delta-weights $\Delta W_k = B_k A_k$ across all $K{=}252$ adapted projections (attention $q,k,v,o$ and MLP $\mathrm{up},\mathrm{gate},\mathrm{down}$ on every decoder layer) and concatenate them into a single gauge-invariant fingerprint $\bm{\theta}$. We then compute the mean pairwise cosine similarity of $\bm{\theta}$ across $N{=}20$ instances on five benchmarks spanning three domains. \Cref{tab:instance_lora} reports the resulting similarity matrix.

\paragraph{Distinct LoRAs per instance.}
The within-dataset similarity ranges from $0.264$ to $0.933$, reflecting the heterogeneity of the underlying instances: Python-coding tasks vary widely in specification, whereas olympiad-style \textsc{minerva-math} problems share a highly templated structure. This spread indicates that \method{} naturally allocates more adapter variation to datasets with diverse instances and less to near-templated ones. This is exactly the behaviour one would expect of an architecture that genuinely conditions on the input rather than collapsing to a fixed per-task adapter.

\paragraph{Domain-aware geometry.}
The off-diagonal entries exhibit clear block structure: within-domain pairs consistently show higher similarity than cross-domain pairs. The math block and the code block are the two highest cross-dataset entries, while the math\,$\times$\, code crossings are the lowest. Notably, the within-dataset similarity of \textsc{mbpp+} nearly coincides with its cross-similarity to \textsc{humaneval+}, which is expected given that both are Python function-level benchmarks with overlapping prompt formats. In all cases the diagonal dominates its row, confirming that $\bm{\theta}$ encodes task semantics faithfully.

\begin{table}[t]
    \centering
    \small
    \setlength{\tabcolsep}{6pt}
    \caption{%
        Mean pairwise cosine similarity of the gauge-invariant LoRA fingerprint
        $\bm{\theta}$ ($N{=}20$ instances per dataset).
        Diagonal: within-dataset; off-diagonal: cross-dataset.
    }
    \label{tab:instance_lora}
    \begin{tabular}{l ccccc}
    \toprule
                       & \textsc{gsm8k} & \textsc{minerva-math} & \textsc{mmlu} & \textsc{mbpp+} & \textsc{humaneval+} \\
    \midrule
    \textsc{gsm8k}         & \textbf{0.519} & 0.267 & 0.148 & 0.133 & 0.094 \\
    \textsc{minerva-math}  &      --          & \textbf{0.933} & 0.215 & 0.187 & 0.220 \\
    \textsc{mmlu}          &      --          &         --       & \textbf{0.424} & 0.063 & 0.099 \\
    \textsc{mbpp+}         &        --        &        --        &     --           & \textbf{0.264} & 0.252 \\
    \textsc{humaneval+}    &     --           &          --      &        --        &           --     & \textbf{0.315} \\
    \bottomrule
    \end{tabular}
\vspace{-0.2in}
\end{table}

\subsection{Ablation Study: Instance-Specificity of Weight Perturbations}
\label{sec:ablation_instance_lora}

A key design choice in \method{} is to generate \emph{instance-specific} perturbations: the LoRA factors injected into the receiver are conditioned on both the sender's reasoning and the particular input.
To validate that this instance-level conditioning is essential to the observed gains, we conduct two ablations that progressively degrade specificity while holding all other variables constant.


\begin{wrapfigure}{r}{0.5\textwidth}
    \vspace{-1.4em}
    \centering
    \includegraphics[width=1\linewidth]{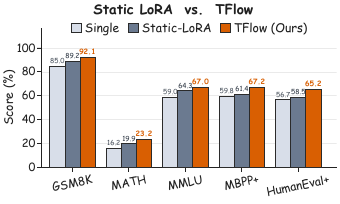} 
    \vspace{-1.8em}
    \caption{\textbf{Static LoRA vs.\ \method{}} performance.}
    \label{tab:static_lora_ablation}
    \vspace{-1.4em}
\end{wrapfigure}

\paragraph{Static LoRA.}
To isolate instance-conditioned parameter generation from standard parameter-efficient adaptation, we replace the parameter generator of TFLOW with a conventional LoRA adapter shared across all inputs, while keeping the backbone, adapted modules, LoRA rank, and training data unchanged.

As shown in Figure~\ref{tab:static_lora_ablation}, Static-LoRA consistently improves over the Single-Agent baseline on all benchmarks. However, TFLOW achieves substantially stronger performance, outperforming Static-LoRA by 4.29 points on average, with especially clear gains on more challenging reasoning and code-oriented benchmarks such as MBPP+ and HumanEval+. This shows that TFLOW’s advantage goes beyond added trainable capacity and stems from input-dependent modulation of the receiver.

\begin{wrapfigure}{r}{0.59\textwidth}
    \vspace{-1.6em}
    \centering
    \includegraphics[width=1\linewidth]{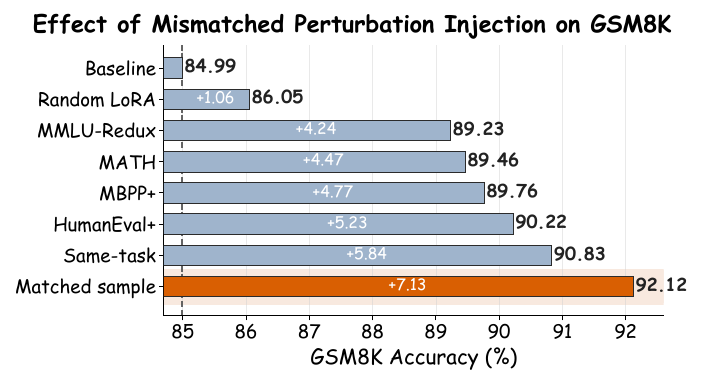} 
    \vspace{-1.5em}
    \caption{\textbf{Mismatched perturbation injection} on GSM8K.}
    \label{fig:mismatch}
    \vspace{-1em}
\end{wrapfigure}

\paragraph{Mismatched Perturbation Injection.}

We further investigate whether TFLOW perturbations encode instance-level information by fixing the receiver input and replacing its matched perturbation with LoRA factors from random sources, other tasks, same-task samples, or the matched sample.

Figure~\ref{fig:mismatch} shows that random LoRA perturbations bring only marginal gains, while cross-task perturbations still improve over the baseline, indicating that TFLOW captures transferable collaborative signals. However, same-task perturbations perform better, and the matched-sample perturbation achieves the highest accuracy, suggesting that TFLOW encodes not only task-level knowledge but also fine-grained instance-specific cues. This confirms that the generated perturbations are meaningfully tied to the target input, rather than serving as generic task-level or adapter-like updates.
\section{Conclusion}

We presented \method{}, a weight-space collaboration paradigm for multi-agent LLM systems. 
Rather than routing inter-agent knowledge through natural language, \method{} maps sender agents' hidden states into query-specific low-rank perturbations of a frozen receiver model. 
These perturbations are fused and transiently injected at inference time, enabling instance-level adaptation without increasing context length or requiring latent-space compatibility.
Across five benchmarks, \method{} consistently outperforms the single-agent baseline and achieves competitive accuracy with text-based multi-agent systems while using substantially fewer tokens and lower latency. 
Our findings indicate that model weights can serve as an effective communication medium among agents, suggesting a promising direction for efficient and scalable multi-agent LLM collaboration.



\bibliography{neurips_2026}
\bibliographystyle{nips}

\newpage

\appendix

\section{Limitations}
\label{sec:limitations}

While \method{} offers an efficient alternative to text-based multi-agent communication, it also has several limitations.

First, the communication channel is less interpretable than natural language. In text-based MAS, intermediate messages can be inspected by humans to understand what each agent contributed. In contrast, \method{} transmits information through low-rank weight perturbations, whose semantic content is difficult to directly interpret. This may complicate debugging, attribution, and safety auditing, especially when the generated perturbations lead to unexpected receiver behavior.

Second, the performance gap on HumanEval+ indicates that \method{} cannot fully recover the benefits that text-based MAS may obtain from increased generation length. Text-based collaboration not only transmits inter-agent information but also expands the overall reasoning and refinement budget through explicit intermediate rationales, plans, and deliberation traces. These additional tokens can directly benefit tasks such as code synthesis, where longer solutions, iterative correction, and detailed implementation planning are often important. \method{} avoids passing sender-written messages to the receiver and therefore removes much of this extra generation overhead. This improves efficiency, but it also means that \method{} may not capture all gains arising from the longer generation process used by TextMAS. Future work may explore hybrid systems that combine concise textual messages with transient parameter perturbations to balance interpretability, generation budget, and efficiency.

\section{Additional Method Details}
\label{app:method_details}

This appendix provides additional details for \method{} that are omitted from the main text for brevity. We describe sender conditioning and layer aggregation, the parameter-generator architecture, sender fusion and transient receiver injection, the training objective, optimization algorithms, and computation cost.

\subsection{Notation}
\label{app:notation}

We use \(N\) to denote the number of agents. Agents \(A_1,\dots,A_{N-1}\) are senders and \(A_N\) is the receiver. All agents share the same frozen backbone parameters \(\btheta\) and are distinguished by role-specific prompts \(p_i\). For a query \(q\), sender \(A_i\) processes the prompted input \((p_i,q)\) and produces hidden states \(\{\bH_i^{(l)}\}_{l=0}^{L}\), where
\[
  \bH_i^{(l)}
  \in
  \R^{T_i\times d}.
\]
Here \(T_i\) is the sender sequence length and \(d\) is the hidden dimension of the frozen backbone.

We use \(L\) for the number of targeted receiver layers and \(M\) for the number of targeted linear modules per targeted layer. The frozen weight of module \(m\) in layer \(l\) is denoted by
\[
  \bW_m^{(l)}
  \in
  \R^{d_{\mathrm{out}}^{(l,m)}
  \times
  d_{\mathrm{in}}^{(l,m)}} .
\]
The LoRA rank is \(r\), and \(\alpha\) is the LoRA scaling coefficient. The parameter generator has hidden dimension \(d_{\mathrm{pg}}\) and \(N_{\mathrm{pg}}\) transformer blocks.

\subsection{Sender Conditioning and Layer Aggregation}
\label{app:conditioning}

For each query, every sender runs one frozen forward pass on its prompted input. We retain hidden states from all backbone layers:
\[
  \{\bH_i^{(l)}\}_{l=0}^{L},
  \qquad
  \bH_i^{(l)}\in\R^{T_i\times d}.
\]
Since different layers encode complementary levels of abstraction, we aggregate them using learnable layer weights. Let \(\lambda_l\) be a scalar assigned to layer \(l\), and let \(\tau>0\) be a temperature. The normalized layer weight is
\begin{equation}
  \rho_l
  =
  \frac{\exp(\lambda_l/\tau)}
  {\sum_{l'=0}^{L}
  \exp(\lambda_{l'}/\tau)} .
  \label{eq:app_layer_weight}
\end{equation}
The sender conditioning signal is then
\begin{equation}
  \bC_i
  =
  \sum_{l=0}^{L}
  \rho_l \bH_i^{(l)}
  \in
  \R^{T_i\times d}.
  \label{eq:app_layer_aggregation}
\end{equation}
This construction lets the generator access information from both shallow and deep sender representations while preserving a fixed conditioning dimension.

Before entering the generator, the conditioning signal is projected to the generator hidden dimension:
\begin{equation}
  \widetilde{\bC}_i
  =
  \mathrm{Proj}_{\mathrm{in}}(\bC_i)
  \in
  \R^{T_i\times d_{\mathrm{pg}}}.
  \label{eq:app_condition_projection}
\end{equation}
The sender backbone remains frozen, and gradients do not update \(\btheta\).

\subsection{Parameter Generator}
\label{app:parameter_generator}

The parameter generator \(\mathcal{H}_{\bpsi}\) maps each sender conditioning signal \(\bC_i\) to LoRA factors for every targeted receiver module. For each layer \(l\in\{1,\dots,L\}\) and module \(m\in\{1,\dots,M\}\), it outputs
\begin{equation}
  \bA_{i,m}^{(l)}
  \in
  \R^{r\times d_{\mathrm{in}}^{(l,m)}},
  \qquad
  \bB_{i,m}^{(l)}
  \in
  \R^{d_{\mathrm{out}}^{(l,m)}\times r}.
  \label{eq:app_lora_shapes}
\end{equation}
Thus, the corresponding low-rank update has shape
\[
  \bB_{i,m}^{(l)}\bA_{i,m}^{(l)}
  \in
  \R^{d_{\mathrm{out}}^{(l,m)}
  \times
  d_{\mathrm{in}}^{(l,m)}}.
\]

\paragraph{Token-grid initialization.}
The generator starts from a learnable token grid
\[
  \bQ_{\mathrm{grid}}
  \in
  \R^{L\times HW\times d_{\mathrm{pg}}},
\]
where \(H=H_A+H_B\) separates tokens used for generating \(\bA\)- and \(\bB\)-factors, and \(W\) indexes module--rank slots. The projected sender condition \(\widetilde{\bC}_i\) initializes this grid through cross-attention:
\begin{equation}
  \bZ_i^{(0)}
  =
  \mathrm{CA}
  \bigl(
    \bQ_{\mathrm{grid}},
    \widetilde{\bC}_i,
    \widetilde{\bC}_i
  \bigr)
  +
  \bE_{\mathrm{layer}}
  +
  \bE_{\mathrm{slot}},
  \label{eq:app_grid_init}
\end{equation}
where \(\bE_{\mathrm{layer}}\) encodes the target receiver-layer index and \(\bE_{\mathrm{slot}}\) encodes module and rank-slot positions.

\paragraph{Multi-axis transformer trunk.}
The initialized grid is refined by \(N_{\mathrm{pg}}\) transformer blocks. Each block contains cross-layer self-attention, intra-layer self-attention, conditioning cross-attention, and a feed-forward network. For block \(t\), we write
\begin{align}
  \bU_i^{(t)}
  &=
  \bZ_i^{(t)}
  +
  \mathrm{SA}_{L}
  \bigl(
    \mathrm{LN}(\bZ_i^{(t)})
  \bigr),
  \label{eq:app_sa_layer}
  \\
  \bV_i^{(t)}
  &=
  \bU_i^{(t)}
  +
  \mathrm{SA}_{HW}
  \bigl(
    \mathrm{LN}(\bU_i^{(t)})
  \bigr),
  \label{eq:app_sa_slot}
  \\
  \bY_i^{(t)}
  &=
  \bV_i^{(t)}
  +
  \mathrm{CA}
  \bigl(
    \mathrm{LN}(\bV_i^{(t)}),
    \widetilde{\bC}_i,
    \widetilde{\bC}_i
  \bigr),
  \label{eq:app_ca_condition}
  \\
  \bZ_i^{(t+1)}
  &=
  \bY_i^{(t)}
  +
  \mathrm{FFN}
  \bigl(
    \mathrm{LN}(\bY_i^{(t)})
  \bigr).
  \label{eq:app_ffn}
\end{align}
Here, \(\mathrm{SA}_{L}\) attends across targeted receiver layers while batching over the \(HW\) axis, and \(\mathrm{SA}_{HW}\) attends within each layer across module--rank slots. The conditioning cross-attention re-grounds the parameter tokens in the sender representations at every block. RoPE is applied along the active attention axis.

\paragraph{Detokenization into LoRA factors.}
After the final generator block, the grid is split along the \(H\) axis into \(\bA\)-tokens and \(\bB\)-tokens:
\[
  \bZ_i^{(N_{\mathrm{pg}})}
  =
  [
    \bZ_{i,A},
    \bZ_{i,B}
  ].
\]
Two linear heads project these tokens into parameter slices:
\begin{equation}
  \bS_{i,A}
  =
  h_A(\bZ_{i,A}),
  \qquad
  \bS_{i,B}
  =
  h_B(\bZ_{i,B}).
  \label{eq:app_detok_heads}
\end{equation}
The slices are rearranged according to layer, module, and rank-slot indices to produce
\[
  \left\{
    \bA_{i,m}^{(l)},
    \bB_{i,m}^{(l)}
  \right\}_{l=1,m=1}^{L,M}.
\]
All generator parameters are shared across senders. Sender-specific LoRA factors arise solely from sender-specific conditioning signals \(\bC_i\).

\subsection{Sender Fusion and Transient Injection}
\label{app:fusion_injection}

When multiple senders are active, \method{} fuses their generated updates at the perturbation level. This is important because fusing \(\bA\)- and \(\bB\)-factors separately would create unintended cross-sender products. Specifically,
\[
  \left(
    \sum_i \gamma_i \bB_i
  \right)
  \left(
    \sum_j \gamma_j \bA_j
  \right)
  =
  \sum_{i,j}
  \gamma_i\gamma_j
  \bB_i\bA_j,
\]
which includes terms \(\bB_i\bA_j\) for \(i\neq j\). We therefore fuse complete low-rank perturbations \(\bB_i\bA_i\).

The sender-fusion gate predicts one scalar score for each sender:
\begin{equation}
  s_i
  =
  g_{\bw}(\bC_i).
  \label{eq:app_gate_score}
\end{equation}
Scores are normalized across active senders:
\begin{equation}
  \gamma_i
  =
  \frac{\exp(s_i)}
  {\sum_{j=1}^{N-1}\exp(s_j)}.
  \label{eq:app_gate_weight}
\end{equation}
The fused transient update for module \(m\) in layer \(l\) is
\begin{equation}
  \Delta\bW_m^{(l)}
  =
  \frac{\alpha}{r}
  \sum_{i=1}^{N-1}
  \gamma_i
  \bB_{i,m}^{(l)}
  \bA_{i,m}^{(l)} .
  \label{eq:app_fused_update}
\end{equation}

For a receiver activation matrix \(\bX\in\R^{T\times d_{\mathrm{in}}^{(l,m)}}\), the patched linear map is
\begin{equation}
  \bX
  \mapsto
  \bX(\bW_m^{(l)})^\top
  +
  \bX(\Delta\bW_m^{(l)})^\top.
  \label{eq:app_patched_linear}
\end{equation}
The dense matrix \(\Delta\bW_m^{(l)}\) need not be explicitly materialized. The additive branch can be computed in low-rank form:
\begin{equation}
  \bX(\Delta\bW_m^{(l)})^\top
  =
  \frac{\alpha}{r}
  \sum_{i=1}^{N-1}
  \gamma_i
  \bigl(
    \bX(\bA_{i,m}^{(l)})^\top
  \bigr)
  (\bB_{i,m}^{(l)})^\top .
  \label{eq:app_low_rank_branch}
\end{equation}
The frozen weight tensor \(\bW_m^{(l)}\) is never overwritten. The update exists only during the current receiver forward pass and is removed immediately afterward.

\subsection{Training Objective}
\label{app:training_objective}

The trainable parameter set is
\[
  \Theta
  =
  \{
    \bpsi,
    \mathrm{Proj}_{\mathrm{in}},
    \{\lambda_l\}_{l=0}^{L_{\mathrm{total}}},
    \bw
  \},
\]
where \(\bpsi\) denotes the parameter-generator weights, \(\mathrm{Proj}_{\mathrm{in}}\) denotes the conditioning projection, \(\{\lambda_l\}\) are the layer-aggregation scalars, and \(\bw\) denotes the fusion-gate parameters. The frozen backbone parameters \(\btheta\) are excluded from \(\Theta\).

\paragraph{Task loss.}
For a training tuple \((x_q,t,\sigma)\), where \(x_q\) is the query-side input, \(t\) is the target completion, and \(\sigma\) is the data-source identifier, the task loss is the masked next-token cross-entropy over completion tokens:
\begin{equation}
  \mathcal{L}_{\mathrm{task}}(\Theta;x_q,t)
  =
  -
  \sum_{u\in\mathcal{I}_{\mathrm{ans}}}
  \log
  p_{\btheta,\Theta}
  \bigl(
    t_u
    \mid
    t_{<u}, x_q
  \bigr),
  \label{eq:app_task_loss}
\end{equation}
where \(\mathcal{I}_{\mathrm{ans}}\) denotes answer-token positions. The distribution \(p_{\btheta,\Theta}\) is computed by the frozen receiver equipped with the transient updates produced by \(\Theta\).

\paragraph{Flattened generated update.}
To define the diversity regularizer, we flatten all fused updates for the current input:
\begin{equation}
  v_{\Theta}(x_q)
  =
  \mathrm{vec}
  \left(
    \left\{
      \Delta\bW_m^{(l)}(x_q)
    \right\}_{l=1,m=1}^{L,M}
  \right).
  \label{eq:app_flatten_lora}
\end{equation}
In implementation, this vector can be represented either by explicitly flattened dense offsets or by an equivalent flattened low-rank representation, as long as the same representation is used consistently for the current vector and the cache.

\paragraph{Cache-based diversity regularizer.}
For each data source \(\sigma\), we maintain a stop-gradient cache vector \(v_{\mathrm{prev}}^{(\sigma)}\), corresponding to the generated update from the previous optimization step involving that source. The diversity regularizer is
\begin{equation}
  \mathcal{L}_{\mathrm{div}}(\Theta;x_q)
  =
  \cos^2
  \bigl(
    v_{\Theta}(x_q),
    v_{\mathrm{prev}}^{(\sigma(x_q))}
  \bigr).
  \label{eq:app_div_loss}
\end{equation}
This objective penalizes both parallel and anti-parallel collapse and is minimized when the current update is orthogonal to the cached update. Since \(v_{\mathrm{prev}}^{(\sigma)}\) is detached, the regularizer does not introduce an additional forward pass.

The total objective is
\begin{equation}
  \mathcal{L}(\Theta)
  =
  \mathbb{E}_{(x_q,t,\sigma)\sim\mathcal{D}}
  \left[
    \mathcal{L}_{\mathrm{task}}(\Theta;x_q,t)
    +
    \lambda_{\mathrm{div}}
    \mathcal{L}_{\mathrm{div}}(\Theta;x_q)
  \right].
  \label{eq:app_total_loss}
\end{equation}

\subsection{Training Algorithm}
\label{app:training_algorithm}

\begin{algorithm}[h]
\caption{Training \method{}}
\label{alg:training}
\begin{algorithmic}[1]
\Require Frozen backbone \(f_{\btheta}\); trainable parameters \(\Theta\); sender prompts \(\{p_i\}_{i=1}^{N-1}\); receiver prompt \(p_N\); dataset \(\mathcal{D}\); diversity weight \(\lambda_{\mathrm{div}}\).
\State Initialize diversity cache \(\{v_{\mathrm{prev}}^{(\sigma)}\}\) for each data source \(\sigma\).
\For{each minibatch \(\mathcal{B}\subset\mathcal{D}\)}
  \State Initialize minibatch loss \(\mathcal{L}_{\mathcal{B}}\leftarrow 0\).
  \For{each example \((x_q,t,\sigma)\in\mathcal{B}\)}
    \For{each sender \(i=1,\dots,N-1\)}
      \State Run frozen sender forward on \((p_i,x_q)\) and collect \(\{\bH_i^{(l)}\}_{l=0}^{L_{\mathrm{total}}}\).
      \State Aggregate hidden states into \(\bC_i\) using \Cref{eq:app_layer_aggregation}.
      \State Generate LoRA factors \(\{\bA_{i,m}^{(l)},\bB_{i,m}^{(l)}\}_{l,m}\leftarrow\mathcal{H}_{\bpsi}(\bC_i)\).
    \EndFor
    \State Compute gate weights \(\gamma_i=\mathrm{softmax}_i(g_{\bw}(\bC_i))\).
    \State Fuse sender updates into \(\Delta\bW_m^{(l)}\) for all targeted modules using \Cref{eq:app_fused_update}.
    \State Temporarily patch the targeted receiver modules.
    \State Run the frozen receiver on \((p_N,x_q,t)\) and compute \(\mathcal{L}_{\mathrm{task}}\) over completion tokens.
    \State Form \(v_{\Theta}(x_q)\) and compute \(\mathcal{L}_{\mathrm{div}}\) using the stop-gradient cache \(v_{\mathrm{prev}}^{(\sigma)}\).
    \State Accumulate \(\mathcal{L}_{\mathcal{B}}\leftarrow\mathcal{L}_{\mathcal{B}}+\mathcal{L}_{\mathrm{task}}+\lambda_{\mathrm{div}}\mathcal{L}_{\mathrm{div}}\).
    \State Remove the temporary receiver patches.
  \EndFor
  \State Update \(\Theta\) using gradients of \(\mathcal{L}_{\mathcal{B}}\).
  \State Update each used cache entry \(v_{\mathrm{prev}}^{(\sigma)}\leftarrow\mathrm{stopgrad}(v_{\Theta}(x_q))\).
\EndFor
\end{algorithmic}
\end{algorithm}

\subsection{Inference Algorithm}
\label{app:inference_algorithm}

At inference time, \method{} follows the same sender-conditioning and parameter-generation process, but omits the task loss, diversity regularizer, and cache update. The generated perturbation is used only during receiver decoding.

\begin{algorithm}[h]
\caption{Inference with \method{}}
\label{alg:inference}
\begin{algorithmic}[1]
\Require Frozen backbone \(f_{\btheta}\); trained parameters \(\Theta\); sender prompts \(\{p_i\}_{i=1}^{N-1}\); receiver prompt \(p_N\); query \(q\).
\For{each sender \(i=1,\dots,N-1\)}
  \State Run frozen sender forward on \((p_i,q)\) and collect \(\{\bH_i^{(l)}\}_{l=0}^{L_{\mathrm{total}}}\).
  \State Aggregate hidden states into \(\bC_i\).
  \State Generate LoRA factors \(\{\bA_{i,m}^{(l)},\bB_{i,m}^{(l)}\}_{l,m}\leftarrow\mathcal{H}_{\bpsi}(\bC_i)\).
\EndFor
\State Compute \(\gamma_i=\mathrm{softmax}_i(g_{\bw}(\bC_i))\).
\State Fuse updates into \(\Delta\bW_m^{(l)}=\frac{\alpha}{r}\sum_{i=1}^{N-1}\gamma_i\bB_{i,m}^{(l)}\bA_{i,m}^{(l)}\).
\State Temporarily patch the targeted receiver modules.
\State Decode the final answer using receiver \(A_N\) conditioned on \((p_N,q)\).
\State Remove all temporary patches.
\end{algorithmic}
\end{algorithm}

\subsection{Computation Cost}
\label{app:computation_cost}

The additional cost of \method{} beyond a standard receiver forward pass comes from sender forwarding, parameter generation, and transient LoRA injection.

\paragraph{Sender forward passes.}
Let \(C_{\mathrm{bb}}(T)\) denote the cost of one frozen backbone forward pass with sequence length \(T\). The \(N-1\) senders incur
\begin{equation}
  C_{\mathrm{sender}}
  =
  \sum_{i=1}^{N-1}
  C_{\mathrm{bb}}(T_i).
  \label{eq:app_sender_cost}
\end{equation}
These sender passes are independent and can be parallelized. They also do not require autoregressive decoding; each sender only produces hidden states for conditioning.

\paragraph{Parameter generation.}
Let \(S=HW\) be the number of parameter tokens per targeted layer. Each generator block applies cross-layer attention, intra-layer attention, conditioning cross-attention, and an FFN. For sender \(i\), the cross-layer attention costs \(O(SL^2d_{\mathrm{pg}})\), the intra-layer attention costs \(O(LS^2d_{\mathrm{pg}})\), the conditioning cross-attention costs \(O(LST_i d_{\mathrm{pg}})\), and the FFN costs \(O(LSd_{\mathrm{pg}}^2)\). Therefore,
\begin{equation}
  C_{\mathrm{gen}}^{(i)}
  =
  O\!\left(
    N_{\mathrm{pg}}
    \left[
      SL^2d_{\mathrm{pg}}
      +
      LS^2d_{\mathrm{pg}}
      +
      LST_i d_{\mathrm{pg}}
      +
      LSd_{\mathrm{pg}}^2
    \right]
  \right),
  \label{eq:app_generator_cost}
\end{equation}
and across all senders,
\begin{equation}
  C_{\mathrm{gen}}
  =
  \sum_{i=1}^{N-1}
  C_{\mathrm{gen}}^{(i)}.
  \label{eq:app_total_generator_cost}
\end{equation}

\paragraph{Transient LoRA injection.}
For a targeted linear module with receiver sequence length \(T\), input dimension \(d_{\mathrm{in}}^{(l,m)}\), output dimension \(d_{\mathrm{out}}^{(l,m)}\), and LoRA rank \(r\), the low-rank branch in \Cref{eq:app_low_rank_branch} costs
\begin{equation}
  O\!\left(
    (N-1)T r
    \left(
      d_{\mathrm{in}}^{(l,m)}
      +
      d_{\mathrm{out}}^{(l,m)}
    \right)
  \right).
  \label{eq:app_lora_module_cost}
\end{equation}
This is small compared with the dense projection cost
\[
  O\!\left(
    T d_{\mathrm{in}}^{(l,m)}
    d_{\mathrm{out}}^{(l,m)}
  \right)
\]
when \(r\ll\min(d_{\mathrm{in}}^{(l,m)},d_{\mathrm{out}}^{(l,m)})\). Summing over all targeted modules gives
\begin{equation}
  C_{\mathrm{inj}}
  =
  O\!\left(
    (N-1)T r
    \sum_{l=1}^{L}
    \sum_{m=1}^{M}
    \left(
      d_{\mathrm{in}}^{(l,m)}
      +
      d_{\mathrm{out}}^{(l,m)}
    \right)
  \right).
  \label{eq:app_total_injection_cost}
\end{equation}

\paragraph{Overall overhead.}
The total additional cost is
\begin{equation}
  C_{\mathrm{extra}}
  =
  C_{\mathrm{sender}}
  +
  C_{\mathrm{gen}}
  +
  C_{\mathrm{inj}}.
  \label{eq:app_total_extra_cost}
\end{equation}
The sender and generator costs scale linearly with the number of active senders \(N-1\), and the injection cost scales linearly with both \(N-1\) and the LoRA rank \(r\). In practice, sender forward passes and per-sender generation are independent and can be batched or parallelized, reducing wall-clock overhead.

\newpage

\end{document}